\documentclass[10pt,journal,compsoc]{IEEEtran}

\usepackage{graphicx}
\usepackage{amsmath}
\usepackage{amssymb}	

\usepackage{multirow}
\usepackage{makecell}
\usepackage{array}
\usepackage{booktabs}	

\usepackage{cite}
\ifCLASSINFOpdf
\else
\fi

\usepackage[ruled,norelsize,vlined,linesnumbered]{algorithm2e}
\usepackage{color}
\makeatletter
\newcommand{\removelatexerror}{\let\@latex@error\@gobble}
\makeatother

\ifCLASSOPTIONcompsoc
  \usepackage[caption=false,font=footnotesize,labelfont=sf,textfont=sf]{subfig}
\else
  \usepackage[caption=false,font=footnotesize]{subfig}
\fi
\begin{document}
	%
	\title{Interpretable System Identification and Long-term Prediction on Time-Series Data}
	%
	%
	%
	%
	
	\author{Xiaoyi Liu,~\IEEEmembership{}
		Duxin Chen$^{*}$,~\IEEEmembership{}
		Wenjia Wei, ~\IEEEmembership{}
		Xia Zhu, ~\IEEEmembership{}
		and Wenwu Yu$^{*}$,~\IEEEmembership{Senior Member,~IEEE}
		
		\thanks{
			* Corresponding author
			
			Xiaoyi Liu, Duxin Chen, and Wenwu Yu are with the Jiangsu Key Laboratory of Networked Collective Intelligence, School of Mathematics, Southeast University, Nanjing 210096, China. (e-mail: xiaoyiliu@seu.edu.cn, chendx@seu.edu.cn, wwyu@seu.edu.cn)
			
	}}
	
	%
	%

\markboth{Journal of \LaTeX\ Class Files,~Vol.~14, No.~8, August~2023}%
{Shell \MakeLowercase{\textit{et al.}}: Bare Advanced Demo of IEEEtran.cls for IEEE Computer Society Journals}
%



\IEEEtitleabstractindextext{%
	\begin{abstract}
		Time-series prediction has drawn considerable attention during the past decades fueled by the emerging advances of deep learning methods. However, most neural network based methods lack interpretability and fail in extracting the hidden mechanism of the targeted physical system. To overcome these shortcomings, an interpretable sparse system identification method without any prior knowledge is proposed in this study. This method adopts the Fourier transform to reduces the irrelevant items in the dictionary matrix, instead of indiscriminate usage of polynomial functions in most system identification methods. It shows an interpretable system representation and greatly reduces computing cost. With the adoption of $l_1$ norm in regularizing the parameter matrix, a sparse description of the system model can be achieved.  Moreover, Three data sets including the water conservancy data, global temperature data and financial data are used to test the performance of the proposed method. Although no prior knowledge was known about the physical background, experimental results show that our method can achieve long-term prediction regardless of the noise and incompleteness in the original data more accurately than the widely-used baseline data-driven methods. This study may provide some insight into time-series prediction investigations, and suggests that an white-box system identification method may extract the easily overlooked yet inherent periodical features and may beat neural-network based black-box methods on long-term prediction tasks.
	\end{abstract}
	
	}

\maketitle

\IEEEdisplaynontitleabstractindextext

%
\IEEEpeerreviewmaketitle

\ifCLASSOPTIONcompsoc
\IEEEraisesectionheading{\section{Introduction}\label{sec:introduction}}
\else
\section{Introduction}
\label{sec:introduction}
\fi

%
%
%
%
\IEEEPARstart{A}{ccurate} time-series prediction is of crucial importance in the era of big data, which is widely used in many fields of modern industry, such as prediction on energy consumption, weather change, disease spreading, traffic flow and economic evolution \cite{Autoformer, traffic, TCN, Informer, time_series_forecast}. It has attracted widespread attention of scientists ranging from mathematics, computer science, and other related specific engineering fields. Therein, a large number of achievements have been obtained and show appreciable performance. However, long (e.g., more than one thousand steps) and stable prediction has still been a tricky but significant problem to untangle due to the lack of deep understanding of the hidden mechanism in the targeted physical system \cite{ultra_long_prediction}. Moreover, most existing methods are proposed in a deep learning framework, which cannot meets the requirement of interpretable and reliable mechanism to be used in large-scale industrial systems. This motivates us to further investigate the mechanism of the system evolution with a white-box format and get rid of the usage of a black-box neural network module.


Towards time-series prediction, two mainstream ways have emerged in order to solve this problem over recent years. One arises from data-driven methods, e.g., neural networks (NN) and statistical learning. These kinds of approaches can form end-to-end prediction without explicit mechanistic model. The early data-driven methods include Back-Propagation Net (BP Net) \cite{BP_Net} and Auto-regressive Integrated Moving Average model (ARIMA) \cite{ARIMA}, which work well in prediction for small-scale system with simple rules. However, in most systems with complex nonlinearity and stochasticity, neither of these models can guarantee prediction accuracy in long-term prediction. In order to increase the capability of long-term prediction, a variety of recurrent neural networks (RNNs) have been proposed. Two widely-used RNNs are Long Short-Term Memory (LSTM) \cite{LSTM} and Gate Recurrent Unit (GRU) \cite{GRU}, which show priority in handling and predicting important events with long interactions in time series \cite{LSTMNet, DeepAR}. But RNNs and their variant models have far more parameters than typical NNs, which require longer time to train the model and have more risk of over-fitting. To overcome this, Temporal Convolution Network (TCN) \cite{TCN} was subsequently proposed. TCN uses causal convolution to model temporal causality, allowing parallel computation of output which can improve network performance compared to RNNs. Unfortunately, using TCN for long-term prediction requires multiple iterations of multi-step (or single-step) forecasting, which will induce error accumulation and thus lose effect in capturing long-term mechanisms \cite{step_predict_problem}. Recently, with the invention of Transformer \cite{Transformer} which only based on self-attention mechanism, many variants of Transformer models \cite{Reformer, Informer, Autoformer} have been successively used for long-term prediction without multiple iteration. Among them, Autoformer \cite{Autoformer} yielded state-of-the-art results with both accuracy and computational speed. Nevertheless, Autoformer requires sufficient amount of data for training, and the prediction length can only reach up to hundreds of time points with guaranteed accuracy. Although these data-driven methods achieved great performance in specific prediction tasks and have been widely used in practical systems, they still suffer from interpretability and stability in extracting physical mechanism. Because they uniquely adopt the black-box NN framework and lack the capability of inferring system dynamics.

The other prevailing way is based on system identification. Considering that most engineering systems evolve with underlying physical mechanisms in real scenes \cite{Sparse_identification_introduction_1, Sparse_identification_introduction_2, Sparse_identification_introduction_3, sparse_graph}, original problem can be transformed into a sparse identification of time-series. Orthogonal Matching Pursuit algorithm (OMP) \cite{OMP} is one of the earliest algorithms to solve this problem. By using greedy idea, this method can quickly obtain the sparse model, but it cannot reach the optimal solution in most cases. To improve it, LASSO \cite{LASSO} and Sparse Bayes Method \cite{SBM} were proposed, respectively, which can obtain the optimal solution under prior knowledge and assumptions. However, such prior assumptions may be difficult to satisfy in reality and the prior knowledge is limited or even unable to obtain in most target systems. Recently, Sparse Identification of Nonlinear Dynamics (SINDy) \cite{SINDy} and Identification of Hybrid Dynamic System (IHYDE) \cite{IHYDE} based on discovering governing potential equation in noisy complex systems showed well performance in real scenes, such as autonomous vehicles, fluid convection, medical applications, etc. Many related models \cite{SINDYc, Implict-SINDy, Abrupt-SINDy, two_step_method} emerged quickly in this field. However, an inevitable bottleneck in current sparse identification methods is the dilemma between balancing the number of basis functions and the complexity of computation. Paying too much attention to the former will lead to unbearable computational time \cite{delemma_1}, which will even spend more time than training large-scale neural networks. On the contrary, taking excessive interest in the latter will lead to inaccurate identification results \cite{SINDy}. Such a dilemma hinders the long-term prediction of sparse identification in time-series.

\begin{figure}[!b]
	\centering
	\includegraphics[width=3.5in]{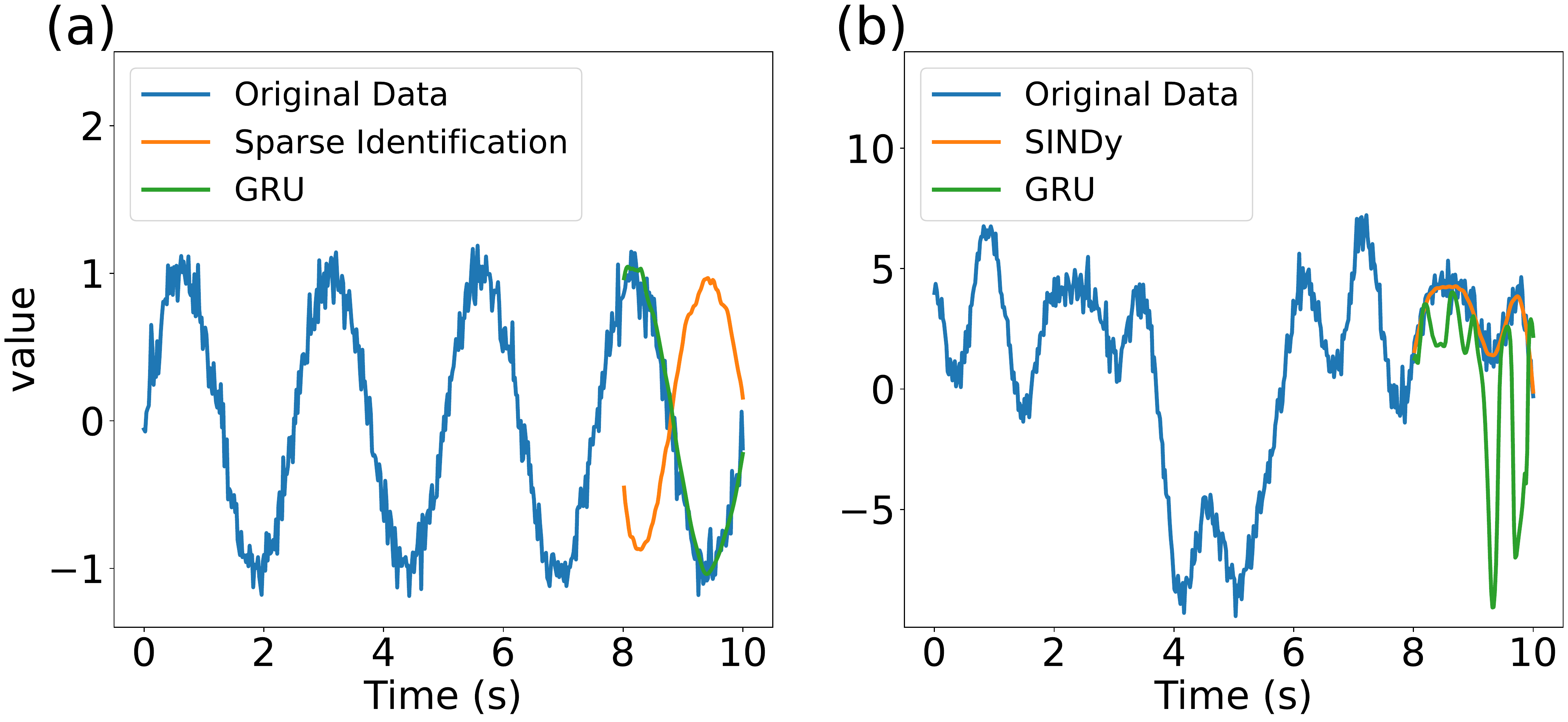}
	\caption{Examples of two prediction methods. (a) Generated rule: $x=\sin(0.5t)+\epsilon$, and figure (b) Generated rule: $x=5\sin(t)+3\cos(2t)-2.3\sin(5t)+1.2\cos(7t)+\epsilon$. The identification function library assumes basis functions as $\sin(t)$, $\cos(t)$, $\sin(2t)$, $\cos(2t)$, ..., $\sin(100t)$, $\cos(100t)$. In both the two cases, the training and prediction duration are 8 and 2 seconds, respectively. }
	\label{problems}
\end{figure}

To further illustrate the current bottlenecks of existing model-based and data-driven methods, a controlled experiment has done. Two classes of time series as follows are constructed. Both model-based and data-driven methods are used for prediction of the constructed time series, respectively. The sparse identification method fails in extracting the mechanism in Fig~\ref{problems}.(a), because the generated rule is not assumed in the basis functions, while GRU predicts well due to the simple generated rule of time series \cite{NN_advantage}. In contrast, when fitting an erratic curve in Fig.~\ref{problems}.(b), NN can only make short-term accurate predictions. But owing to the assumed basis functions including part of the generated rule, SINDy yields better prediction results \cite{SINDy}.

Thus, in order to improve accuracy and computation speed for time-series prediction, especially in long-term tasks, this study proposes a novel sparse system identification method with adaptive basis functions (SIABF), including three implementing steps. In the first step, the prior basis function library is given and the selection criterion is evaluated. Considering that the vast majority of time-series potentially contain periodicity \cite{time_series_period}, Discrete Fourier Transform (DFT) \cite{DFT} is used at first to transfer time domain to frequency domain. By sorting the frequency rank in descending order according to the amplitude of the DFT spectrum, the resorting diagram which implies the strength of intrinsic potential periodicity were obtained. The prior period obtained in the top ranks of the resorting diagram by time-frequency conversion formula \cite{period-frequency} will be used in the prior Fourier basis functions. In the second step, an evaluation index for the quality of prior basis is provided before making the prediction. We define the quasi-periodic index $I^{(10)}$ as the ratio of the top ten fastest decline slope to the maximum amplitude in the sorting diagram, which indicates the period complexity of the whole time series. The value of $I^{(10)}$ can provide an effective reference for the applicability of prior basis selection. The third step is long-term prediction by using sparse identification. The format of the prior basis functions are assumed to mainly include a series of sine and cosine functions with that evaluated top frequency ranks in the first step. With adopting the $l_{1}$ norm in regularizing the parameter matrix \cite{LASSO, SBM}, a sparse identification of time-series is achieved. One may use the identification result to make long-term prediction for time series.

Compared with existing model-based and data-driven methods, SIABF has the following advantages. The prior periods obtained by DFT not only provide the most likely candidates in the basis function library, but also endow the entire model with interpretability in the frequency domain which NN based methods cannot provide \cite{NN_without_interablilty}. SIABF greatly reduces the dimension of basis functions matrix compared to other model-based methods \cite{SINDy, IHYDE, two_step_method}. Moreover, it can achieves long-term prediction by leveraging the underlying dominant frequency hidden in any physical systems, compared to most widely-used celebrated data-driven methods \cite{Autoformer, GRU, LSTMNet}. 


The main contributions of this paper are as follows:

\begin{enumerate}
	\item[1.] A novel interpretable sparse system identification method is proposed, which captures the essential periodic factors with transformation into the frequency domain. The underlying frequency features obtained and used in the basis functions clearly describe the evolution mechanism of the physical system. 
	\item[2.] The proposed SIABF model achieves highly accurate and stable prediction results with less computation and training time in the long-term prediction task on time series.
	\item[3.] The state-of-the-art results are obtained via comprehensive comparisons with other well-accepted latest model-based and data-driven baseline methods under the test on temperature, water conservancy and finance datasets. Obvious priority has been observed in the comparison experiments.
\end{enumerate}

%


The rest of this paper is organized as follows. Section~2 describe the problem and preliminaries of sparse identification for time-series prediction. Section~3 presents the details of the proposed SIABF model. Section~4 shows the comparison with other baseline methods via numerical experiments. Section~5 concludes this paper.

\section{Problem and Preliminaries}
\indent\indent
In this section, the description of sparse identification problem is presented along with some preliminaries in order to describe more clearly the methodology later.

\subsection{Problem Description}
\indent\indent
The traditional sparse identification problem of time-series prediction can be described as follows \cite{SINDy}:
\begin{equation}\label{constraint}
	\mathbf{X}_{k} = \mathbf{\Theta}(\mathbf{t}, \mathbf{\widetilde{X}}) \mathbf {\Xi}_k + \mathcal{N}, k = 1,2,...,m,
\end{equation}
where, $\mathbf{X}_k\in {{\mathbf{R}}^{n}}$, is the state of the $k$-th of $m$ output of the system with $n$ predicted steps. $\mathbf{\Theta}$ is a $p$-dimensional vector-valued function, which represents the basis function library in sparse identification. $\mathbf{t}\in {{\mathbf{R}}^{n}}$ is the prediction time, $\mathbf{\widetilde{X}}\in {{\mathbf{R}}^{(m\times l)\times n}}$ represents the 1, 2, ..., $l$-step data of all systems before the prediction data in the $l$-order dynamic equation. Especially, for the 0-order dynamic equation (i.e., algebraic equation), it has $\mathbf{\widetilde{X}} \equiv 0$. $\mathcal{N}$ is the tolerable error term of the model, and $\mathcal{N} \sim N\left( 0,\sigma^2 \mathbf{I} \right)$. $\mathbf{\Xi}_k$ is the sparse coefficient matrix which is the core of sparse identification. The target is to obtain the sparse matrix $\mathbf{\Xi}_k$ in the model, i.e.,
\begin{align}\label{target}
	\min{\left\| \mathbf{\Xi}_k  \right\|}_{0}=\sum\limits_{i=1}^{(ml+1) p}{{{\left( \xi _{i}^{0} \right)}_{k}}},k = 1,2,...,m.
\end{align}
Here, $0^0 = 0$. The basic idea of time-series prediction with sparse identification is to train $\mathbf{\Xi}_k $ with $\mathbf{X}_k, \mathbf{\widetilde{X}}, \mathbf{\Theta}$ in the sample data, so that $\mathbf{\Xi}_k $ conforms to Eq.~\eqref{target}, and $\mathbf{X}_k, \mathbf{\widetilde{X}}, \mathbf{\Theta}$ are subject to Eq.~\eqref{constraint}.

\subsection{Preliminaries for System Identification}
\indent\indent
Without loss of generality, for sparse identification of time series, two basic problems need to be solved. One is the selection of basis function library $\mathbf{\Theta}$, the other is the applicability of the sparse solution $\mathbf{\Xi}_k$ under equation Eq.~\eqref{constraint} in sample data.

The selection of basis function is a thorny problem in sparse identification. Most of the existing methods put all the basic elementary functions into the library matrix \cite{SINDy, two_step_method}, including polynomial basis, trigonometric basis, logarithmic basis, exponential basis, constant basis, etc. In brief, $\mathbf{\Theta} (\centerdot )$ can be written as
\begin{align}\label{theta}
	\mathbf{\Theta} (\centerdot )=\left[ \begin{matrix}
		| & | & |  & | & |   \\
		\sin /\cos ({c}_{1} \centerdot) & {\centerdot }^{{{c}_{2}}} & e^{(\centerdot )} & \ln (|\centerdot|) & \mathbf{1}  \\
		| & | & |  & | & |   \\
	\end{matrix} \right],
\end{align}
where, $c_1, c_2$ are real numbers and $(\centerdot)$ represents $(\mathbf{t}, \mathbf{\widetilde{X}} _j)$. In the basis function library, because the functions of different exponents or logarithms are linearly correlated, only one exponential or logarithmic basis is required. In this case, the main part of $\mathbf{\Theta} (\centerdot)$ lies in the polynomial and trigonometric basis. Considering that $c_{1}, c_{2}$ in Eq.~\eqref{theta} can be any real number, one solution is to select one basis according to the data and traverse $c_{1}$ or  $c_{2}$ to get as many natural numbers as possible. The theoretical basis for this is Weierstrass' two approximation theorems\cite{Weierstrass}.  This seems to be a good way if one knows the possible composition of the right term of the dynamical equations. Many studies have obtained significant performances \cite{SINDYc, Implict-SINDy, IHYDE} with knowledge of its possible underlying governing equations. However, it is not possible to obtain the form of the right-hand term in advance. Because without a prior knowledge, even fitting the simplest curve with a large number of basis functions, one may get bad prediction results (see Fig.~\ref{problems}.(a)). Meanwhile, excessive selection of basis functions may not only bring the risk of over fitting, but also need to bear large computational pressure, which is called dimensional disaster \cite{dimensional_disaster}. In this circumstance, a novel method to find the basis function library is urgent to be proposed.

The applicability of the sparse solution $\mathbf{\Xi}_k$ is also a tough problem.  Since the original problem is an NP-hard \cite{NP_hard} one, most of the existing solutions are approximate or optimal solutions under some additional conditions. At present, there are three main ways to solve $\mathbf{\Xi}_{k}$ sparsely. The first idea is to solve through greedy thought. OMP \cite{OMP}, Relaxed-Greedy Algorithm (RGA) \cite{RGA} and SINDy \cite{SINDy} are all representatives of such an idea. They select the local optimal solution in each iteration, so that the original NP-hard problem can be solved in polynomial time. Taking SINDy \cite{SINDy} as an example, when solving $\mathbf{\Xi}_{k}$ in the $j$-th iteration, the iterative process is as follows,
\begin{align}\label{SINDy}
	\underset{\mathbf{\Xi}_k }{\mathop{\arg \min }}\,{{\left\| \mathbf{X}_k^{(j)}-\sum\limits_{i\in \Gamma ^ {(j)} }{\theta_{i} (\centerdot )\left(\xi _{i}^{(j)}\right)_k} \right\|}^{2}},
\end{align}
where, $\theta_{i}(\centerdot )$ is the $i$-th column of $\mathbf{\Theta}(\mathbf{t}, \mathbf{\widetilde{X}})$, and the ${\Gamma }^{(0)}$ and ${\Gamma }^{(j+1)}$ are defined as
\begin{align}\notag
	&{{\Gamma }^{(0)}}=\left\{ 1, 2, \cdots ,(ml+1)p \right\},\\
	&{{\Gamma }^{(j+1)}}\leftarrow {{\Gamma }^{(j)}}\bigcap {{\left\{ i\left| |\xi _{i}^{(j)}|\le \varepsilon  \right. \right\}}^{c}}, \notag
\end{align}
where the threshold $\varepsilon$ is the super parameter that can be adjusted. This method can quickly find a sparse solution $\mathbf{\Xi}_k$ by using the least squares method in Eq.~\eqref{SINDy}, but it is obviously not the sparsest case in the model.  

The second idea is to replace the $l_{0}$ norm in Eq.~\eqref{target} with $l_{q}(q\ge1)$ norm, which transforms the original problem into a convex optimization problem. In that case, it can be solved in polynomial time. The original problem is equivalent to the following optimization, which can be described as
\begin{align}\label{LASSO}
	\underset{\mathbf{\Xi}_k }{\mathop{\arg \min }}\,{{\left\| {{X}_{k}}-\mathbf{\Theta}(\mathbf{t}, \mathbf{\widetilde{X}}) \mathbf {\Xi}_k \right\|}^{2}}+\lambda \left\| {{\mathbf{\Xi}}_{k}} \right\|_{{{l}_{q}}}^{q},
\end{align}
where $\lambda$ is the super parameter that can be adjusted. The original problem refers to LASSO with $q=1$, and Ridge Regression with $q=2$. Coordinate descent method\cite{LASSO} and least angle regression\cite{Least_angle_regression} are both mature solutions of the method. Among $l_p$ norm, the best sparse effect can be obtained with $l_1$ norm. However, although the obtained result is sparse, it is not the solution of the original problem. $\mathbf{\Xi}_{k}$ obtained by this method is a good alternative to the original problem in cases where the sparsity of $\mathbf{\Xi}_{k}$ is not strictly required. 

The third idea is to give a prior probability to the coefficient of original problem, and use Bayesian statistical learning method to solve the problem. One may assume that each $\xi_{i}$ is independent of each other, and they all obey the Gaussian distribution $N(0, \sigma_{i}^2)$, i.e.,
\begin{align}\notag
	p({{\xi }_{i}})=\int{N({{\xi }_{i}};0,\sigma _{k}^{2})p(\sigma _{k}^{2})\text{d}\sigma _{k}^{2}},
\end{align}
with $p(\sigma _{k}^{2})\propto 1/\sigma _{k}^{2}$ (Normal-Jeffreys prior). According to Bayesian formula, the posterior probability of $\mathbf{\Xi}_{k}$ is obtained after simplification, i.e.,
\begin{align}
	\begin{split}\notag
		p({{\mathbf{\Xi} }_{k}})\propto -&\mathbf{\Xi} _{k}^{\top }{{H}^{\top }}H{{\mathbf{\Xi} }_{k}}\\
		&-2{{\mathbf{\Xi}}_{k}}{{H}^{\top }}{\mathbf{X}_{k}}-\mathbf{\Xi}_{k}^{\top }\Upsilon (\sigma _{\Xi }^{2}){\mathbf{\Xi}_{k}},
	\end{split}
\end{align}
where $H$ is the kernel function of $\Theta(\centerdot)$, while $\sigma _{\Xi }^{2} = \text{diag}(\sigma_1^{-2},\cdots,\sigma_{(ml+1)p}^{-2})$. The final optimization form can be obtained by using the log maximum likelihood estimation for a posterior probability $p({{\mathbf{\Xi} }_{k}})$, i.e.,
\begin{align}\label{SBL}
	\begin{split}
		\underset{\mathbf {\Xi}_k}{\mathop{\arg \min }}\,\sum\limits_{i=1}^{n}&{\frac{1}{2\sigma _{i}^{2}}{{\left| {\mathbf{X}_{k}^{(i)}}-\mathbf{\Theta}(\mathbf{t}^{(i)}, \mathbf{\widetilde{X}}^{(i)}) \mathbf {\Xi}_k^{(i)}  \right|}}}\\
		&+\frac{n}{2}\sum\limits_{i=1}^{(ml+1)p}{\log \sum\limits_{j=1}^{n}{\left(\xi _{i}^{(j)}\right)_k}},
	\end{split}
\end{align}
where the superscript $i$ (or $j$) represents the value of the $i$-th (or $j$-th) moment in time series. Eq.~\eqref{SBL} is a solvable convex optimization problem. Here, the prior probability does not have to be Gaussian prior, and other types of prior probability are also emerging quickly \cite{improved_SBL_1, improved_SBL_2, improved_SBL_3}. These three ideas Eq.~\eqref{SINDy}-\eqref{SBL} greatly enrich the solution of sparse identification. 

However, following the research line, the naturally inspired question is whether it is possible to choose a model-based rather than a data-driven method to solve the practical prediction problem? To put it another way, can we know the effectiveness of sparse identification methods before solving $\mathbf{\Xi}_k$? The existing methods do not give an index to evaluate the effectiveness of the model, so we can only constantly try various methods including model-based and data-driven, which is quite laborious and time-consuming.

In the following section, we will give a unique and general prior knowledge for the basis functions and propose the evaluation index for SIABF. This will alleviate these two problems to a certain extent.


\section{Sparse Identification with Adaptive Basis Functions (SIABF)}
\begin{figure*}[t]
	\centering
	\includegraphics[width=1\textwidth]{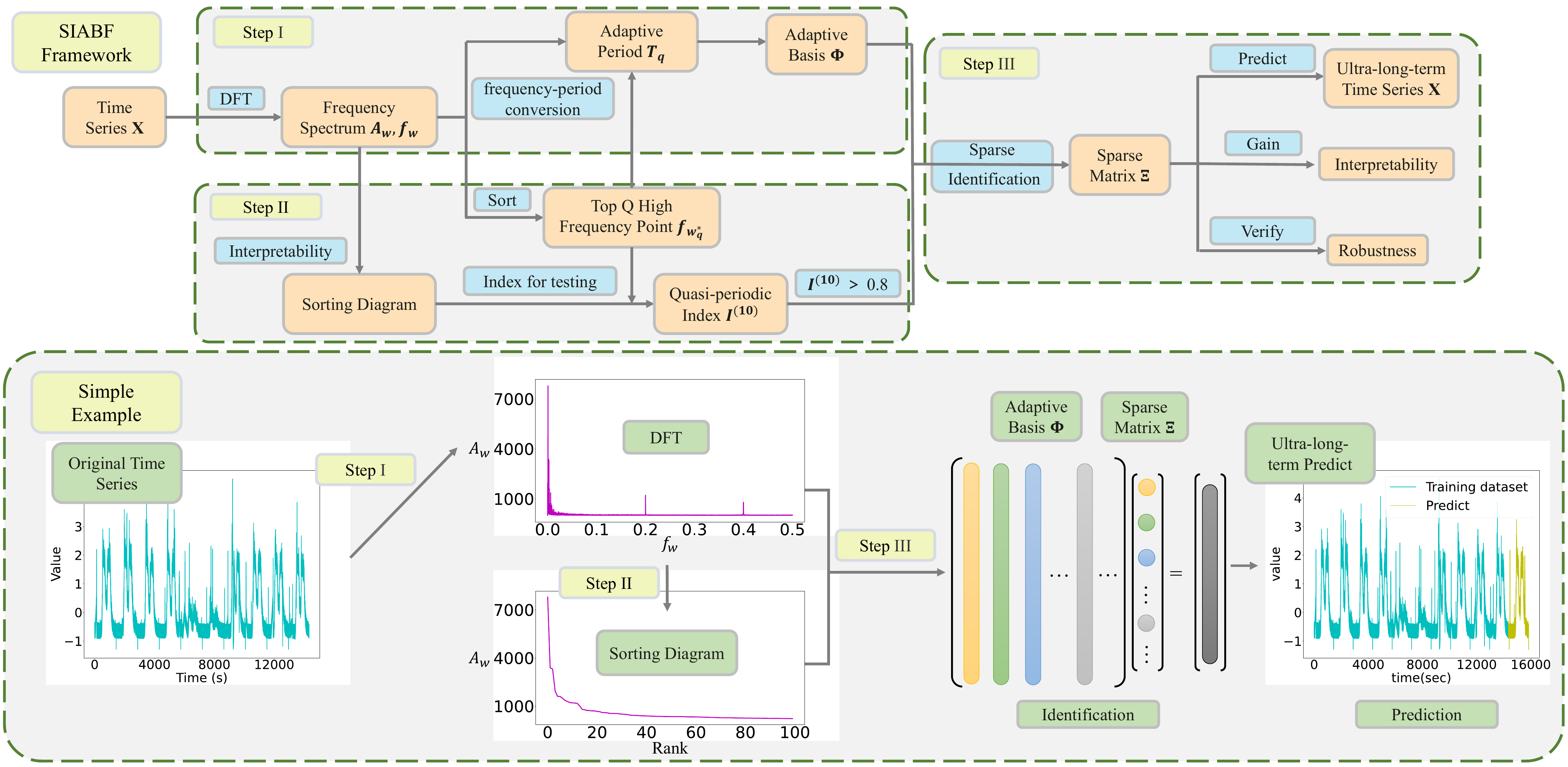} 
	\caption{The framework of SIABF and a simple example.}
	\label{process}
\end{figure*}

\indent\indent
In this section, the core methodology will be introduced. SIABF is divided into three steps. The first step is used to find adaptive basis in time series. The second step is to discriminate the effectiveness of basis. The third step is to compute sparse matrix $\mathbf{\Xi}_k$ and verify the robustness of the model. The implementing process of SIABF is shown in Fig.~\ref{process}. With SIABF, the problems of current model-based method mentioned in section~2 can be effectively solved.

\subsection{Methodology}
\indent\indent
In this part, a method for selecting adaptive basis functions will be proposed, which differs from traditional model-based methods under artificial experience. On this circumstance, a validation method on the effectiveness of adaptive basis is given as well.
\subsubsection{Step \uppercase\expandafter{\romannumeral1}: Adaptive Period}
\indent\indent
First, it is necessary to determine what basis function to select for identification in Eq.~\eqref{theta}. Note that most practical data embed periodicity, tendency and stochasticity\cite{time_series_period}. Targeted prior basis functions should be designed. Considering that there often exist mutation points in the practical sample data and the tail of high-order polynomial identification will produce a large Runge phenomenon\cite{Runge}, polynomial fitting is inadequate. Moreover, owing to the polynomial function has infinite values at infinity, the long prediction points may produce large errors. Thus, the identification basis function library will mainly adopt the triangular base in Eq.~\eqref{theta}. 

For the time series $\mathbf{X}$ of a certain physical system with length $N$, in order to obtain the potential period information hidden in the data, DFT($\mathbf{X}$) with module of amplitude ${{A}_{w}}$ and frequency ${{f}_{w}}$ is adopted,

\begin{equation}\label{Aw}
	{{A}_{w}}=\left| \sum\limits_{s=0}^{N-1}{{X_{s}}\cdot {{e}^{-\frac{j2\pi ws}{N}}}} \right|, w = 0,1,\cdots,\left[\frac{N-1}{2}\right],
\end{equation}
\begin{equation}\label{fw}
	{{f}_{w}}=\frac{w}{N\cdot d},\text{ }
	w = 0,1,\cdots,\left[\frac{N-1}{2}\right],
\end{equation}
where $d$ represents sample frequency, $X_{s}$ refers to the value of the $s$-th component in $\mathbf{X}$ and $j$ is the imaginary unit. Due to the conjugate symmetry of DFT, only positive frequency needs to be take into consideration, which means that $w$ only needs to traverse half length of the data. According to the property of DFT, the greater value of $A_w$, the more influence of the corresponding frequency $f_w$ in the time series data. Therefore, we sort $A_w$ from large to small, selecting top $Q$, and find the corresponding argument $w_{q}^{*}$, i.e., $$w_{q}^{*}={{\underset{w}{\mathop{\arg \max }}\,}^{(q)}}{{A}_{w}}, q=1,2,\cdots,Q,$$
where, $\max^{(q)}$ represents the $q$-th largest number in  $A_w$. In this way, one can find the high frequency $f_{w_q^*}$ of the time series. By using the frequency-period conversion formula ${{T}^{(q)}}=1/f_{w^{*}_{q}}$, the most likely period existing in the time series can be obtained. These periods are recorded as $T^{(1)}, T^{(2)}, \cdots ,T^{(Q)}$ and named as the adaptive period. These ${T}^{(q)}$ are brought into the basis function library of the identification process, in which, each $c_1^{(i)}$ in Eq.~\eqref{theta} can be written as
\begin{align}\label{prior_basis}
	c_{1}^{(q)}=\frac{2\pi}{{T}^{(q)}}, \text{   } q=1,2,\cdots ,Q.
\end{align}
It is worth noting that the change of time series can be basically expressed by these $Q$ periods. Therefore, one can describe the main period of transformation in this time series by selecting the maximum corresponding amplitude in the $Q$ periods. This improves interpretability of the model to a higher level compared with artificially selecting adaptive basis functions or only using polynomials inspired by Taylor expansion. In addition, the number of prior cycles (i.e., the size of $Q$) for selection may not be very large. Compared with the traditional traversal scheme for natural numbers\cite{SINDy, IHYDE}, our model will greatly reduce the computational complexity. This makes the model more suitable for solving large-scale and long-term prediction.

In this way, the basis function in sparse identification is adaptively selected from \eqref{Aw} and \eqref{fw}. It is worth noting that the main parts of adaptive basis functions are chosen only through sample time series, which is adaptive and includes the key periodicity features that are easily ignored. The first problem of sparse identification mentioned above, i.e., selecting adaptive basis functions is solved accordingly.

\subsubsection{Step \uppercase\expandafter{\romannumeral2}: Quasi-periodic Index}

\indent\indent
Generally, complex systems encrypts specific physical mechanism that cannot be entirely uncovered and described. In other words, it is indeed that not all the prior knowledge including periods obtained by DFT can be taken into the identification model for prediction. In this paper, the index $I^{(10)}$ to evaluate the quality of basis function selection will be introduced before identification and prediction. The significance of this index is to avoid wasting time training unnecessary models. Moreover, this index also provides some insight into distinguishing the boundaries of model-based and data-driven models.

The original time series need to be represented by as few adaptive basis functions as possible. One advantage is that for a small number of basis function identification, the probability of forming over-fitting is very low. On the contrary, if the time series need a large number of adaptive periods to form the basis function, even if the original data is restored, it will not be used for prediction due to the high probability of over-fitting.

\begin{figure}[!h]
	\centering
	\includegraphics[width=1\columnwidth]{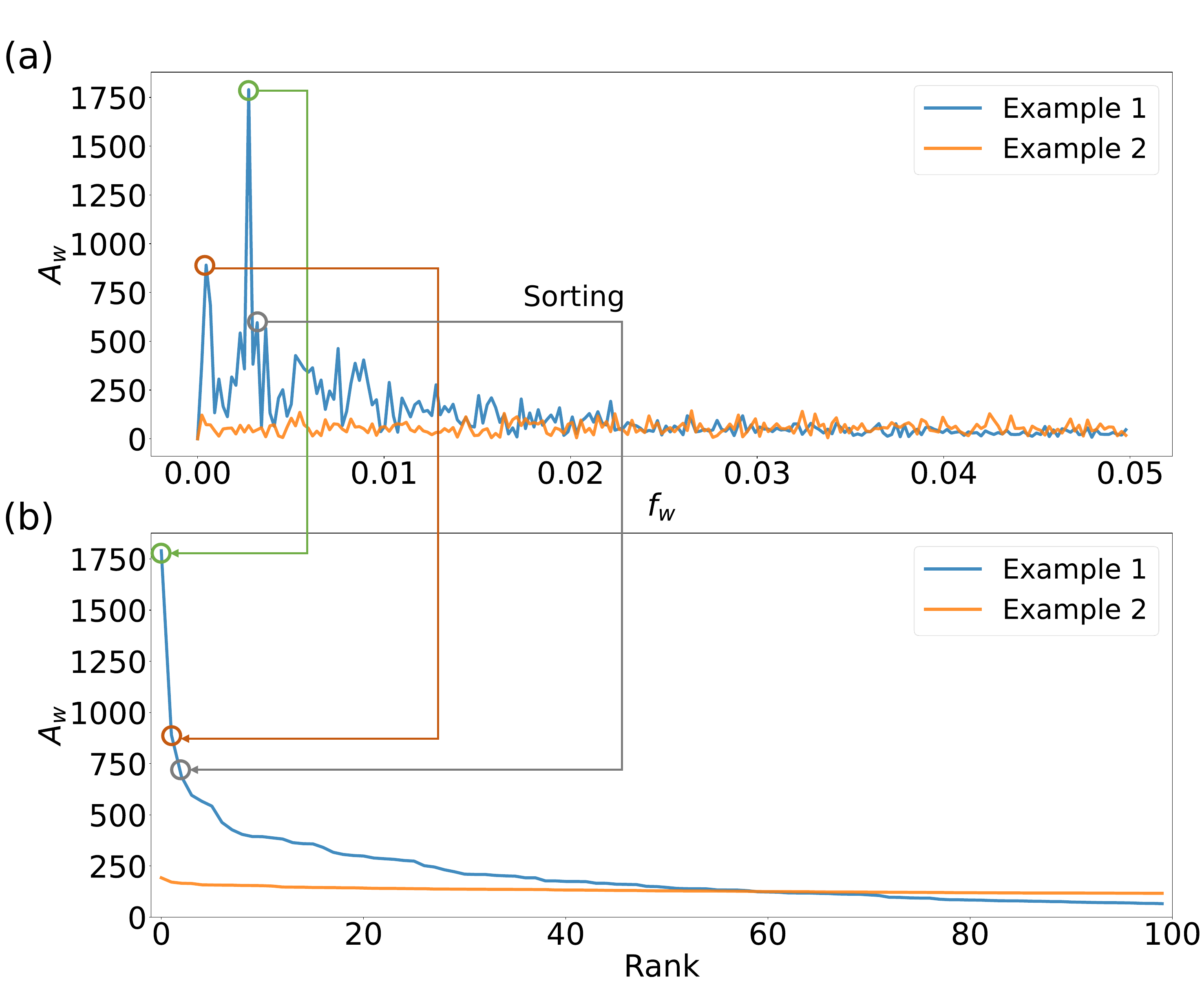} 
	\caption{Two examples of DFT(a) and sorting diagram(b).}
	\label{quasi}
\end{figure}

In order to quantify the effectiveness of the proposed method on adaptive period selection, the amplitude in DFT will be sorted to obtain the sorting diagram (Fig.~\ref{quasi}). According to the property of DFT, the ordinate in sorting diagram represents the amplitude at the corresponding frequency. If the amplitude at the top ranks decreases fast enough in the diagram, that is, the modulus of most DFT results approaches zero, it can be concluded that the period feature affecting the time series is only limited to the top few ranks of frequency with larger amplitudes. Thus, it is sufficient to only use a few prior periods to describe the time series.

Accordingly, the quasi-periodic index $I^{(10)}$ is proposed, which is defined as the ratio of the sum of the top ten slopes $k_{\max}^{(i)}$ in any sorting diagram formed by $Q$ prior basis functions with the maximum amplitude in the DFT, i.e.,
\begin{align}\notag
	{{I}^{(10)}}={\sum\limits_{i=1}^{10}{k_{\max }^{(i)}}}\bigg/{\max {{A}_{w}}}.
\end{align}
Quasi-periodic index, $I^{(10)}$, represents the contribution of the ten most important frequencies to the overall time series. We can empirically judge whether the obtained adaptive period can be further used for sparse identification according to the value of $I^{(10)}$. From the current experimental situation, typically, if $I^{(10)} \ge 0.8$ (Fig.~\ref{quasi}  {Example 1}), the effect in long-term prediction will be much better than NN based methods, but if $I^{(10)} \le 0.5$ (Fig.~\ref{quasi} {Example 2}), such a system identification method may not work. When $I^{(10)}$ is between the middle value ($0.5<I^{(10)}<0.8$), it has its own pros and cons compared with the data-driven method. Note that, the larger value of $I^{(10)}$, the greater possibility that this method is better than data driven-methods in long-term prediction. This indicator may provide an effective criterion for distinguishing model-based or data-driven methods.

In this way, one can judge the applicability of this method before sparse identification. The second problem of sparse identification mentioned above is solved.

\subsubsection{Step \uppercase\expandafter{\romannumeral3: Sparse Identification and Robustness Test}}
\indent\indent
In this step, $\mathbf{\Xi}_{k}$ can be identified sparsely by any traditional identification method with the obtained prior basis in step~\uppercase\expandafter{\romannumeral1}-\uppercase\expandafter{\romannumeral2}. 

Similar to Eq.~\eqref{SINDy}-\eqref{SBL}, $l_1$ norm is used to identify the sparse matrix $\mathbf{\Xi}_{k}$ with the prior basis function, i.e.,
\begin{align}\label{solve}
	\underset{{\mathbf{\Xi}_{k}}}{\mathop{\arg \min }}\,\int_{0}^{T}{\left( {{\left\| {\mathbf{X}_{k}}-\mathbf\Theta (\mathbf{t},{\tilde{\mathbf{X}}}){\mathbf{\Xi }_{k}} \right\|}^{2}} \right)\text{d}\mathbf{t}}+\lambda {{\left\| {\mathbf{\Xi }_{k}} \right\|}_{1}},
\end{align}
where $T$ is the identification duration and $\lambda$ is a hyperparameter, which regulates the sparsity of $\mathbf{\Xi}_{k}$. The value of $\lambda$ can be obtained by the cross validation method\cite{LASSO, two_step_method}. The solution process of Eq.~\eqref{solve} is to solve a convex optimization problem, and the sparse solution can be quickly obtained through Coordinate descent method \cite{LASSO} or least angle regression \cite{Least_angle_regression}.

The resultant of $\mathbf{\Xi}_{k}$ captures the truly effective adaptive basis function in time series. These identified basis functions, including their coefficients, will be used together for long-term prediction. Each step can be expressed as,
\begin{align}\label{predict}
	\mathbf{X}_{k}^{(t+1)}=\mathbf\Theta (\mathbf{t}+1,{{\tilde{\mathbf{X}}}}){{\mathbf\Xi }_{k}}, k=1,2,\cdots,m.
\end{align}
Moreover, in algebraic equation, Eq.~\eqref{predict} does not need iteration. In this case, it will have an extreme long prediction ability without error accumulation.

Furthermore, it is necessary to test the robustness of SIABF. The following experiment is conducted: in the control group, the complete training data are used to predict through SIABF. For the experimental group, 5\% of the training data are deleted. For the deleted data, the missing values are interpolated, so as to keep the dimension of the training data consistent with the control group. Secondly, a random disturbance is given to the training data, which does not necessarily follow the normal distribution, but is completely random. In this case, the training data will be different from the control group. The experiments are carried out on the experimental group and the control group, and the prediction indicators are compared, respectively. If the prediction effect of the experimental group is not significantly weaker than that of the control group, this shows that SIABF does have robustness.

\subsection{Implementation and Algorithm}

\indent\indent
Adding the adaptive period and quasi-periodic index $I^{(10)}$ to the identification methods, by using $l_1$ norm as a penalty term and considering noise to the time series, a complete SIABF model process is obtained. SIABF can find the basis function library adaptively, as well as achieve self-evaluation. 

As shown in Fig.~\ref{process}, the entire identification process can be divided into two phrases with three steps. The first and second step can obtain the adaptive basis with verification. The third step solves sparse identification problem and completes the prediction. Recalling the entire model, the proposed method is summarized in Algorithm~1.
\begin{figure}[!h]
	\removelatexerror
	\begin{algorithm}[H]
		\KwIn{Time-series $\mathbf{X}_{k}, k=0, 1, \cdots, m$.}
		\KwOut{Sparse martrix $\mathbf{\Xi}_{k}$ and long-term predicted time series $\mathbf{\hat{X}}_{k}$, $k=0, 1, \cdots, m.$}
		/* Step \uppercase\expandafter{\romannumeral1} \& \uppercase\expandafter{\romannumeral2}: */\\
		Initialization and standardization.\\
		\For{$k=0, 1, \cdots, m$}{
			${A}_{w}^{k}, {f}_{w}^{k} = \text{DFT}(X_k)$.\\
			Select the $q$ largest from ${A}_{w}^{k}$, and obtain the frequency $f_{w^*}^{k}$ corresponding to the amplitude.\\
			Compute prior basis function $\mathbf{\Theta} (\centerdot )$ with frequency-period conversion and Eq.~\eqref{prior_basis}.\\
			\eIf{$I^{(10)}_k>0.8$}{Having a high degree of confidence to make accurate long-term prediction.}{Method comparison is required.}
		}
		/* Step \uppercase\expandafter{\romannumeral3}: */\\
		Compute coefficient matrix $\mathbf{\Xi}_k$ with Eq.~\eqref{solve}.\\
		Solve the long-term prediction with Eq.~\eqref{predict}.\\
		Test the robustness of SIABF.
		\caption{Spare Identification with Adaptive Basis Functions (SIABF)}
	\end{algorithm}
\end{figure}

\section{Experiments}
\subsection{Experimental Settings}
\subsubsection{Datasets}

\indent\indent
To test the performance and stability of the proposed model SIABF, three time series datasets are used, including water conservancy, global temperature and finance dataset\cite{water_conservancy, temperature, Finance_data}.

\textbf{Water conservancy dataset (Dataset 1)}, from NOAA National Water Model Retrospective dataset, records the mean, maximum and standard deviation of water conservancy data of a river with an sampling interval of 1 hour. The recording time is as long as several decades, and the number of the recording points is more than ten thousands.

\textbf{Global temperature dataset (Dataset 2)}, from NCEP-NCAR Reanalysis, records the global temperature for nearly 70 years since 1950. The recording spatial resolution is $1.875^\circ\times1.875^\circ$. The number of spatio-temporal data points exceeds 500 thousands.

\textbf{Finance dataset (Dataset 3)} records the financial data of 22 major cities in China. This dataset includes nine days records with sampling interval of 1 hour. The total data volume exceeds 100 thousands. 

\subsubsection{Baselines}

\indent\indent
In order to demonstrate the effectiveness of SIABF, several commonly and newly used time series prediction models are selected. The baselines are classified into two categories based on model-based and data-driven methods. 

The compared model-based methods include:
\begin{itemize}
	\item  \textbf{ARIMA} \cite{ARIMA} Auto-regressive Integrated Moving Average model. It is generally recorded as ARIMA$(p, d, q)$, where $p$ is the number of auto-regressive terms, $q$ is the number of moving average terms, and $d$ is the number of differences (orders) to make it a stationary sequence.
	\item  \textbf{SBL} \cite{SBM} Sparse Bayesian Learning method. A common and mature sparse identification method with the assumption that prior distribution of coefficients follows the normal distribution of zero mean.
	\item  \textbf{SINDy} \cite{SINDy} A method for discovering governing equations by the threshold least square. This method can quickly find the sparse solution of the coefficient matrix by using the greedy idea.
\end{itemize}

The compared data-driven methods include:
\begin{itemize}
	\item  \textbf{LSTM}\cite{LSTM} A commonly used RNN for time series prediction is specially designed to solve the long-term dependence of the general RNN.
	\item  \textbf{GRU}\cite{GRU} An improved RNN for LSTM. The network can achieve a prediction result close to that of LSTM with fewer training parameters.
	\item  \textbf{ResNet}\cite{ResNet} 2016 ImageNet champion network. After that, the network has been applied in various scenarios including time series.
	\item  \textbf{Autoformer}\cite{Autoformer} The state-of-art network for long-term prediction at present. It has achieved the best results in the prediction of multiple open datasets.
\end{itemize}

\subsubsection{Evaluation Methodology}

\indent\indent
For fair comparison, all the experiments were performed on the same machine and all the data are normalized. The prediction results are judged from the perspectives of time, accuracy and interpretability. In terms of accuracy, four popular indicators are adopted including RMSE (Root Mean Squared Error), MAE(Mean Absolute Error), R$^2$(R square) and MAPE(Median Absolute Percentage Error) to judge their performances. In order to prevent the influence of the near zero point on MAPE, different from the traditional MAPE definition, the following format is adopted
\begin{align}\notag
	\text{MAPE} = \text{Median}\left\{ \frac{\left| {{y}_{1}}-{{{\hat{y}}}_{1}} \right|}{\left| {{y}_{1}} \right|+\varepsilon },\frac{\left| {{y}_{2}}-{{{\hat{y}}}_{2}} \right|}{\left| {{y}_{2}} \right|+\varepsilon },\cdots ,\frac{\left| {{y}_{n}}-{{{\hat{y}}}_{n}} \right|}{\left| {{y}_{n}} \right|+\varepsilon } \right\},
\end{align}
where $\hat{y}$ is the prediction value and $\varepsilon$ is a small positive number in case the denominator is zero. Each method runs ten times to obtain the average results.

\subsubsection{Parameter settings}
\indent\indent
Optimal parameters suitable for validation data as the initial input of all baselines have been searched and tested. After the repeated tests, the hyperparameters of model-based methods are set as: $\lambda$ = 5e-4 in LASSO, $\alpha_1=\lambda_1$ = 1e-8, $\alpha_2=\lambda_2$ = 5e-5  in SBL, $\epsilon$ = 1e-2 in SINDy. $b, p, q$ are automatically selected in ARIMA. Parts of hyperparameters in data-driven methods are set as: batch size = 50, learning rate = 1e-3, all the epochs in training of network $\ge$ 200.

\subsection{Performance Comparison}
\begin{table*}[!t]
	\centering
	\renewcommand\thetable{\Roman{table}}
	\roman{table}
	\caption{Performance of different models on three datasets. All the experiments consider long-term prediction. The predicted lengths of three datasets are 1460, 1220, 1440 points, respectively.}
	\begin{tabular}{ccccccccccc}
		\hline 
		\specialrule{0em}{1pt}{1pt}
		\multicolumn{2}{c}{Predict Type} & \multicolumn{4}{c}{Complete training dataset}& \multicolumn{4}{c}{Incomplete and noisy training dataset}&\multirow{2}{*}[-2ex]{\makecell{Elapsed\\Time (s)}}\\ \cmidrule(r){3-6} \cmidrule(r){7-10}
		\multicolumn{2}{c}{Metric} &  RMSE  &  MAE  &  R$^2$  &  \makecell{MAPE\\(\%)}&  RMSE  &  MAE  &  R$^2$  &  \makecell{MAPE\\(\%)} & \\  \hline\hline
		\specialrule{0em}{1pt}{1pt}
		\multirow{7}{*}[-0.5ex]{\makecell{Data\\-set 1}}
		
		& ARIMA & 1.052 & 0.859 & -1.37E5 & 35.011 & 1.052 & 0.861 & -7.89E5 & 34.517 & 30.874\\
		& SBL & 0.791 & 0.548 & -0.833 & 23.153 & 0.789 & 0.547 & -0.831 & 23.148 & 0.042 \\
		& SINDy & 0.864 & 0.643 & 0.046 & 38.186 & 0.864 & 0.643 & 0.050 & 38.323 & \textbf{0.022} \\
		& LSTM & 1.058 & 0.876 & -9.22E3 & 33.054  & 1.586 & 1.271 & -2.009 & 72.813 & 718.962 \\
		& GRU & 1.330 & 1.113 & -4.459 & 32.536 & 1.427 & 1.078 & -1.736 & 34.386 & 650.110 \\
		& ResNet & 1.087 & 0.800 & -3.243 & 44.520 & 1.097 & 0.806 & -2.183 & 48.667 & 71.883 \\
		& Autoformer & 0.814 & 0.613 & -1.382 & 25.668 & 0.903 & 0.661 & -1.817 & 26.782 & 92.802 \\
		& \textbf{SIABF} & \textbf{0.491} & \textbf{0.393} & \textbf{0.777} & \textbf{19.222}  & \textbf{0.494} & \textbf{0.392} & \textbf{0.772} & \textbf{19.749} & 0.247 \\
		
		\hline\hline
		\specialrule{0em}{1pt}{1pt}
		\multirow{7}{*}[-0.5ex]{\makecell{Data\\-set 2}}
		
		& ARIMA & 1.067 & 0.928 & -600.733 & 0.401& 1.063 & 0.925 & -513.806 & 0.401 & 56.346 \\
		& SBL & 0.575 & 0.476 & 0.590 & 0.206& 0.572 & 0.473 & 0.593 & 0.205 & \textbf{0.003} \\
		& SINDy & 0.599 & 0.491 & 0.562 & 0.213& 0.595 & 0.489 & 0.565 & 0.212 & 0.008 \\
		& LSTM & 0.602 & 0.474 & 0.604 & 0.205& 0.709 & 0.567 & 0.444 & 0.245 & 482.761 \\
		& GRU & 0.683 & 0.553 & 0.509 & 0.239& 0.666 & 0.537 & 0.508 & 0.232 & 337.221 \\
		& ResNet & 0.585 & 0.471 & 0.532 & 0.204& 0.529 & 0.420 & 0.646 & 0.182 & 38.537 \\
		&Autoformer   &0.766   &0.629   &0.005    &0.271    &0.727   &0.589   &-0.148    &0.254    &183.429  \\
		& \textbf{SIABF} & \textbf{0.430} & \textbf{0.334} & \textbf{0.790} & \textbf{0.144}& \textbf{0.421} & \textbf{0.328} & \textbf{0.794} & \textbf{0.142} & 0.033 \\
		
		\hline\hline
		\specialrule{0em}{1pt}{1pt}
		\multirow{7}{*}[-0.5ex]{\makecell{Data\\-set 3}}
		
		& ARIMA & 0.999 & 0.747 & -16.372 & 123.700& 1.134 & 0.822 & -39.582 & 257.555 & 176.488 \\
		& SBL & 0.901 & 0.822 & 0.070 & 64.558& 0.898 & 0.819 & 0.036 & 64.522 & \textbf{0.003} \\
		& SINDy & 1.207 & 1.124 & -10.855 & 84.089& 1.202 & 1.120 & -10.858 & 83.714 & 0.008 \\
		& LSTM & 0.638 & 0.447 & 0.541 & 46.816& 0.610 & 0.426 & 0.416 & 80.708 & 1495.221 \\
		& GRU & 0.540 & 0.388 & 0.328 & 50.072& 0.595 & 0.435 & 0.240 & 60.308 & 1025.846 \\
		& ResNet & 0.561 & 0.410 & 0.151 & 63.778& 0.647 & 0.484 & 0.118 & 101.053 & 124.194 \\
		&Autoformer   &1.065   &0.909   &-3.273    &174.858    &1.147   &0.979   &-3.112    &144.069        &141.836\\
		& \textbf{SIABF} & \textbf{0.336} & \textbf{0.240} & \textbf{0.863} & \textbf{41.269}& \textbf{0.345} & \textbf{0.250} & \textbf{0.855} & \textbf{49.225} & 0.032 \\
		
		\hline
	\end{tabular}
	\label{compare}    
\end{table*}
\indent\indent
The performance comparison results is shown in Table.~\ref{compare}. For each dataset, two types of prediction are considered: Complete and uncompleted training data for long-term prediction. For each type, total eight methods including the proposed SIABF are tested, and five indicators on accuracy and elapsed time for evaluation are used. From the
results in Table.~\ref{compare}, the following observations are found.

Firstly, SIABF achieves the consistent state-of-the-art results in both complete and incomplete training data. Compared to data-driven methods, SIABF gives 40 \% RMSE improvement (0.491$\rightarrow$0.814) in water conservancy data, 26\% improvement in temperature (0.430$\rightarrow$0.585), 38\% improvement in fiance (0.336$\rightarrow$0.540). Overall, SIABF yields a 33 \% average RMSE improvement among baseline data-driven methods.

Secondly, in all prediction methods, only R$^ 2$ of SIABF is constantly greater than 0.5. In most methods, R$^ 2$ is less than zero in long-term prediction. It clearly shows that most current methods can not accurately predict the trend of future data, that is, they cannot accurately predict the peak and valley values of the data. However, SIABF can accurately grasp long-term trend of the data, which can be attributed to two reasons: 1) SIABF can accurately calculate the potential periods underlying the data, although no prior knowledge of the data was known. 2) SIABF can sparsely select the potential prior periods to ensure that there is no over-fitting phenomenon in the training set.

Thirdly, SIABF saves a lot of training time compared with the data-driven methods. In this paper, although all data-driven methods are trained in the highly configured GPU, training time of each experiment is as long as several minutes. In contrast, SIABF can make long-term prediction within one second maintaining the prediction accuracy which is hundreds or thousands of times faster than the NN based methods.

Finally, SIABF is robust to incomplete and noisy data. In the second type of experiments, 5\% of the training data are randomly deleted, and irregular noise is added to all the training data. However, it is obvious that SIABF is not affected by incomplete training data or noise. Almost all the indicators in Table.~\ref{compare} change less than the baseline methods.

To sum up, SIABF is better than current methods in terms of precision, speed and robustness in the long-term prediction experiments of time series.

\subsection{Interpretability}
\indent\indent
Another advantage of model based-methods is that they have a good interpretability compared with the NN based data-driven methods. Water conservancy dataset is adopted as an example to show the interpretability of SIABF.

\subsubsection{Period implies prior information}
\indent\indent
Unlike most model-based methods, e.g., \cite{SINDy, IHYDE, two_step_method}, the possible adaptive basis functions are unknown for the datasets in our experiments. However, some interpretable information can be provided through the periodic prior method. The sorting diagram for the top ten ranks of period in water conservancy dataset is shown in Fig.~\ref{prior} (right).

The periods most likely to exist in the data can be found through SIABF. Carefully checking these obtained prior periods, it is surprising to find that they are highly interpretable compared to neural networks. In the sorting diagram, the prior period corresponding to the top three ranks are 365, 2920, 1460, respectively, corresponding to the duration of one, eight and four years, respectively. This implies that water conservancy has periodic changes in one, four and eight years. Relevant professional research also confirms this observation \cite{water_period}. Similarly, other cycle lengths can be divided by 16 years. The impact of these prior periods on water conservancy data can be analyzed as well by the observed amplitudes.

Through this example, it is clear that SIABF also provides some prior knowledge even if  the potential functions underlying the data are unknown. It shows that SIABF can efficiently select the basis functions, and indicates that the parameters in SIABF are more interpretable than those in neural networks.

\subsubsection{Quasi-periodic index work}
\begin{figure}[!h]
	\centering
	\includegraphics[width=3.5in]{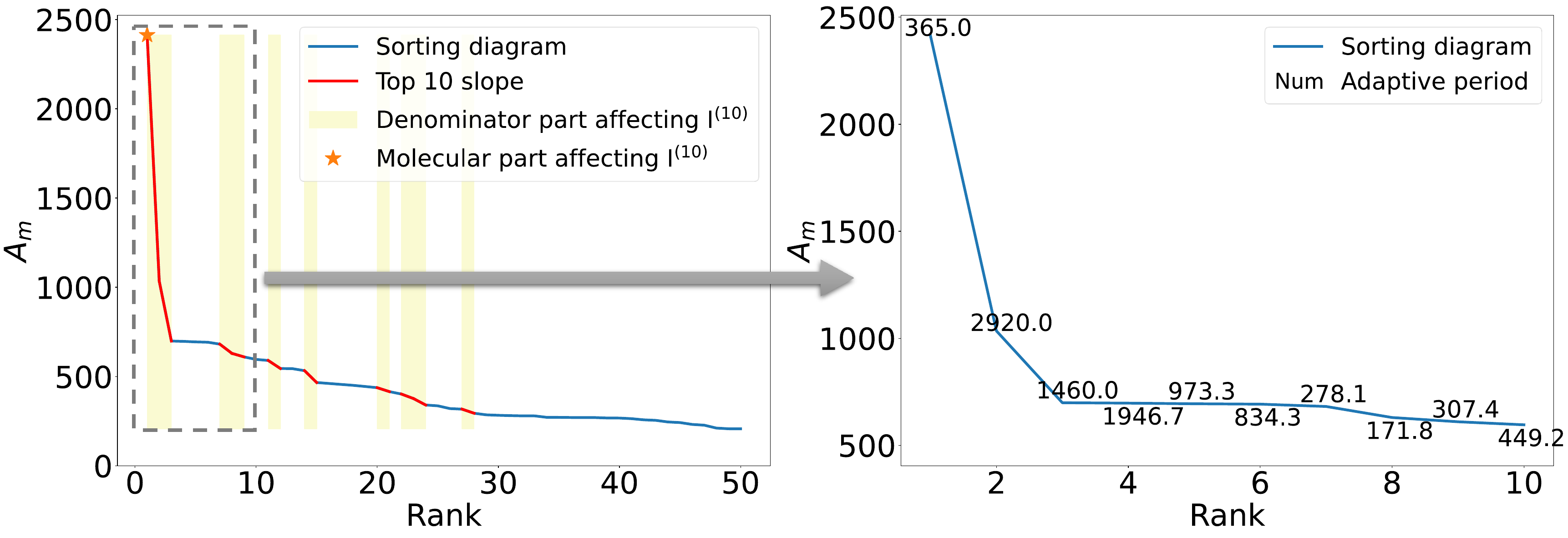}
	\caption{Sorting diagram for calculating $I^{(10)}$ (left) and the visualization of top ten tank adaptive period (right).}
	\label{prior}
\end{figure}

\indent\indent
Furthermore, the quasi-periodic index $I^{(10)}$ is calculated in the water conservancy dataset as an example. With $Q=50$, and the calculation process of quasi-periodic index is shown in Fig.~\ref{prior} (left).

According to Eq.~\eqref{quasi}, $I^{(10)}$ in this dataset is $0.830 > 0.8$. This shows that the time series formed by water conservancy data can be expressed with fewer prior periods, although it is not obvious intuitively (Fig.~\ref{appendix}). In this case, the model-based method is basically better than most data-driven methods. This can also be verified in Table.~\ref{compare}. In this circumstance, SIABF provides a considerable criterion to distinguish between model-based and data-driven methods.

\section{Conclusion}
\indent\indent
In this paper, a novel model-based method, SIABF, for long-term prediction on time series with no prior knowledge is proposed. The adaptive period obtained by simply implementing DFT shows a highly-convincing format of underlying basis functions. The proposed quasi-periodic index, $I^{(10)}$, also provides a high confidence in quantifying the prediction effects before adopting model-based white-box or data-driven black-box methods. When the quasi-periodic index is relatively large (typically $\ge0.8$), SIABF has a great probability to obtain excellent prediction performance.

Moreover, SIABF provides an interpretable, efficient and effective identification and prediction framework with a simple implementing process. It is worth noting that this study indicates that underlying periodicity cannot be overlooked in any data, since it plays a crucial role in quantifying the physical mechanism of complex systems. In the future, more intrinsic features beyond periodicity will be taken into consideration in the framework, and SIABF will be extended to identifying and predicting networked systems.


%

\appendices
\section{Visualization and Code}
\begin{figure}[!h]
	\centering
	\includegraphics[width=3.5in]{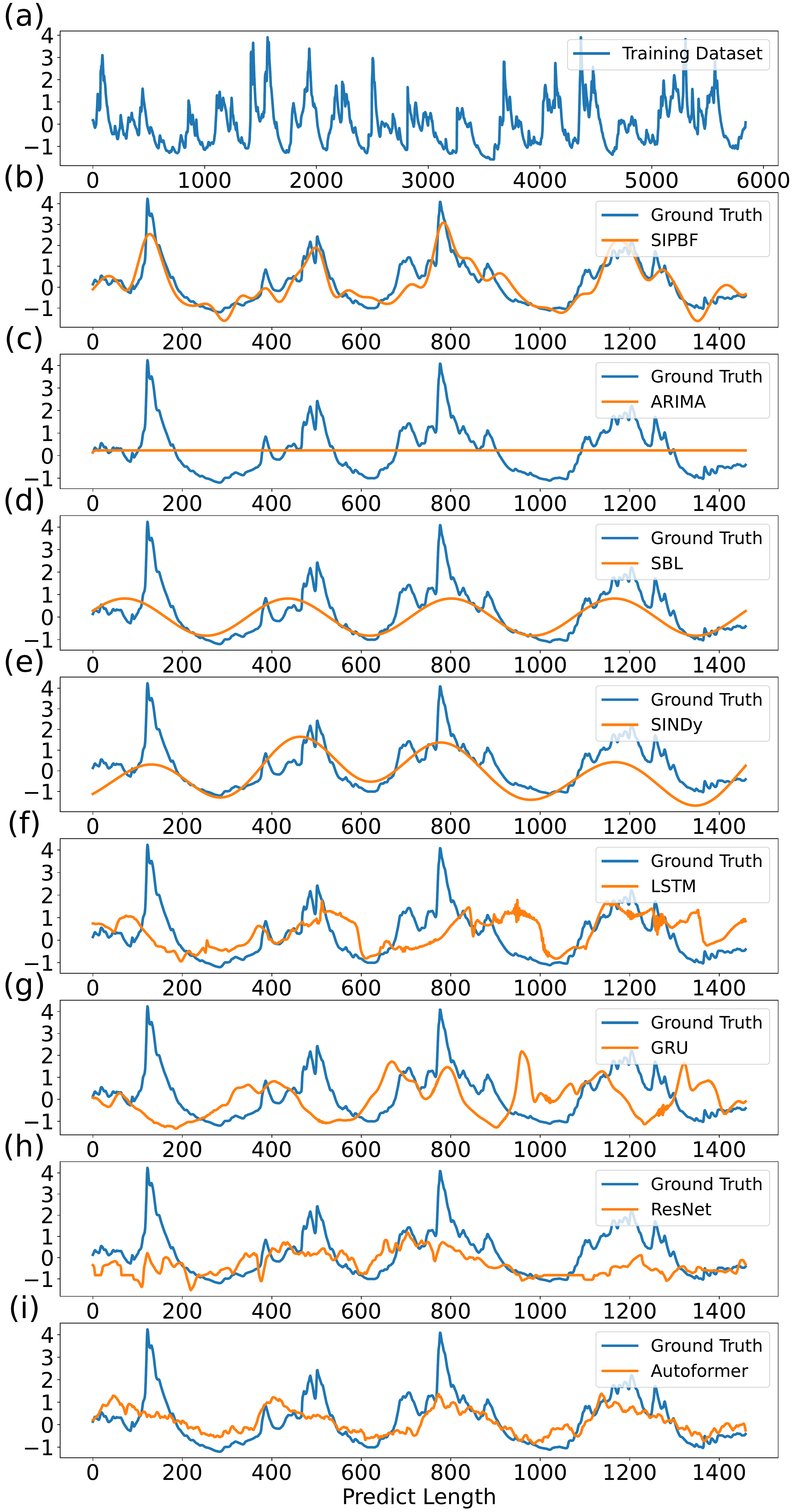}
	\caption{Different methods are applied to the visualization of long-term prediction of water conservancy data. (a) is training dataset, and (b)-(i) show the visualization of test dataset of SIABF, ARIMA, SBL, SINDy, LSTM, GRU, ResNet and Autoformer, respectively.}
	\label{appendix}
\end{figure}

\indent\indent	
Taking water conservancy data as an example, the visualization of long-term prediction of different models are shown in Fig.~\ref{appendix}. Herein, the training data length is 16 years (data points = 365*16), while the prediction length is 4 years (data points = 365*4).

Moreover, our code is available at \cite{Finance_data}.

\ifCLASSOPTIONcaptionsoff
\newpage
\fi



%

%
\bibliographystyle{IEEEtran}
\bibliography{reference}

\begin{thebibliography}{10}
\providecommand{\url}[1]{#1}
\csname url@samestyle\endcsname
\providecommand{\newblock}{\relax}
\providecommand{\bibinfo}[2]{#2}
\providecommand{\BIBentrySTDinterwordspacing}{\spaceskip=0pt\relax}
\providecommand{\BIBentryALTinterwordstretchfactor}{4}
\providecommand{\BIBentryALTinterwordspacing}{\spaceskip=\fontdimen2\font plus
\BIBentryALTinterwordstretchfactor\fontdimen3\font minus
  \fontdimen4\font\relax}
\providecommand{\BIBforeignlanguage}[2]{{%
\expandafter\ifx\csname l@#1\endcsname\relax
\typeout{** WARNING: IEEEtran.bst: No hyphenation pattern has been}%
\typeout{** loaded for the language `#1'. Using the pattern for}%
\typeout{** the default language instead.}%
\else
\language=\csname l@#1\endcsname
\fi
#2}}
\providecommand{\BIBdecl}{\relax}
\BIBdecl

\bibitem{Autoformer}
H.~Wu, J.~Xu, J.~Wang, and M.~Long, ``Autoformer: Decomposition transformers
  with auto-correlation for long-term series forecasting,'' \emph{Advances in
  Neural Information Processing Systems}, vol.~34, pp. 22\,419--22\,430, 2021.

\bibitem{traffic}
Z.~Zhou, Y.~Wang, X.~Xie, L.~Chen, and C.~Zhu, ``Foresee urban sparse traffic
  accidents: A spatiotemporal multi-granularity perspective,'' \emph{IEEE
  Transactions on Knowledge and Data Engineering}, 2020.

\bibitem{TCN}
A.~Borovykh, S.~Bohte, and C.~W. Oosterlee, ``Conditional time series
  forecasting with convolutional neural networks,'' \emph{arXiv preprint
  arXiv:1703.04691}, 2017.

\bibitem{Informer}
H.~Zhou, S.~Zhang, J.~Peng, S.~Zhang, J.~Li, H.~Xiong, and W.~Zhang,
  ``Informer: Beyond efficient transformer for long sequence time-series
  forecasting,'' in \emph{Proceedings of the AAAI Conference on Artificial
  Intelligence}, vol.~35, no.~12, 2021, pp. 11\,106--11\,115.

\bibitem{time_series_forecast}
J.~Deng, X.~Chen, R.~Jiang, X.~Song, and I.~W. Tsang, ``A multi-view multi-task
  learning framework for multi-variate time series forecasting,'' \emph{IEEE
  Transactions on Knowledge and Data Engineering}, 2022.

\bibitem{ultra_long_prediction}
X.~Xu, L.~Zhang, Q.~Kong, C.~Gui, and X.~Zhang, ``Enhanced-historical average
  for long-term prediction,'' in \emph{2022 2nd International Conference on
  Computer, Control and Robotics (ICCCR)}.\hskip 1em plus 0.5em minus
  0.4em\relax IEEE, 2022, pp. 115--119.

\bibitem{BP_Net}
D.~E. Rumelhart, G.~E. Hinton, and R.~J. Williams, ``Learning representations
  by back-propagating errors,'' \emph{Nature}, vol. 323, no. 6088, pp.
  533--536, 1986.

\bibitem{ARIMA}
R.~H. Shumway and D.~S. Stoffer, ``Arima models,'' in \emph{Time Series
  Analysis and Its Applications}.\hskip 1em plus 0.5em minus 0.4em\relax
  Springer, 2017, pp. 75--163.

\bibitem{LSTM}
S.~Hochreiter and J.~Schmidhuber, ``Long short-term memory,'' \emph{Neural
  Computation}, vol.~9, no.~8, pp. 1735--1780, 1997.

\bibitem{GRU}
J.~Chung, C.~Gulcehre, K.~Cho, and Y.~Bengio, ``Empirical evaluation of gated
  recurrent neural networks on sequence modeling,'' \emph{arXiv preprint
  arXiv:1412.3555}, 2014.

\bibitem{LSTMNet}
G.~Lai, W.-C. Chang, Y.~Yang, and H.~Liu, ``Modeling long and short-term
  temporal patterns with deep neural networks,'' in \emph{The 41st
  international ACM SIGIR conference on research \& development in information
  retrieval}, 2018, pp. 95--104.

\bibitem{DeepAR}
D.~Salinas, V.~Flunkert, J.~Gasthaus, and T.~Januschowski, ``Deepar:
  probabilistic forecasting with autoregressive recurrent networks,''
  \emph{International Journal of Forecasting}, vol.~36, no.~3, pp. 1181--1191,
  2020.

\bibitem{step_predict_problem}
B.~Alhnaity, S.~Kollias, G.~Leontidis, S.~Jiang, B.~Schamp, and S.~Pearson,
  ``An autoencoder wavelet based deep neural network with attention mechanism
  for multi-step prediction of plant growth,'' \emph{Information Sciences},
  vol. 560, pp. 35--50, 2021.

\bibitem{Transformer}
A.~Vaswani, N.~Shazeer, N.~Parmar, J.~Uszkoreit, L.~Jones, A.~N. Gomez,
  {\L}.~Kaiser, and I.~Polosukhin, ``Attention is all you need,''
  \emph{Advances in neural information processing systems}, vol.~30, 2017.

\bibitem{Reformer}
N.~Kitaev, {\L}.~Kaiser, and A.~Levskaya, ``Reformer: The efficient
  transformer,'' \emph{arXiv preprint arXiv:2001.04451}, 2020.

\bibitem{Sparse_identification_introduction_1}
Y.~Gu, J.~Jin, and S.~Mei, ``$ l_0 $ norm constraint lms algorithm for sparse
  system identification,'' \emph{IEEE Signal Processing Letters}, vol.~16,
  no.~9, pp. 774--777, 2009.

\bibitem{Sparse_identification_introduction_2}
R.~Wang, C.~Zhang, J.~Bian, Z.~Wang, F.~Nie, and X.~Li, ``Sparse and flexible
  projections for unsupervised feature selection,'' \emph{IEEE Transactions on
  Knowledge and Data Engineering}, 2022.

\bibitem{Sparse_identification_introduction_3}
X.~Ma, Y.~Wang, X.~Chu, L.~Ma, W.~Tang, J.~Zhao, Y.~Yuan, and G.~Wang,
  ``Patient health representation learning via correlational sparse prior of
  medical features,'' \emph{IEEE Transactions on Knowledge and Data
  Engineering}, 2022.

\bibitem{sparse_graph}
Y.~Ye and S.~Ji, ``Sparse graph attention networks,'' \emph{IEEE Transactions
  on Knowledge and Data Engineering}, 2021.

\bibitem{OMP}
Y.~C. Pati, R.~Rezaiifar, and P.~S. Krishnaprasad, ``Orthogonal matching
  pursuit: Recursive function approximation with applications to wavelet
  decomposition,'' in \emph{Proceedings of 27th Asilomar conference on signals,
  systems and computers}.\hskip 1em plus 0.5em minus 0.4em\relax IEEE, 1993,
  pp. 40--44.

\bibitem{LASSO}
R.~Tibshirani, ``Regression shrinkage and selection via the lasso,''
  \emph{Journal of the Royal Statistical Society. Series B, Statistical
  Methodology}, vol.~73, no.~3, 2011.

\bibitem{SBM}
M.~E. Tipping, ``Sparse bayesian learning and the relevance vector machine,''
  \emph{Journal of machine learning research}, vol.~1, no. Jun, pp. 211--244,
  2001.

\bibitem{SINDy}
S.~L. Brunton, J.~L. Proctor, and J.~N. Kutz, ``Discovering governing equations
  from data by sparse identification of nonlinear dynamical systems,''
  \emph{Proceedings of the national academy of sciences}, vol. 113, no.~15, pp.
  3932--3937, 2016.

\bibitem{IHYDE}
Y.~Yuan, X.~Tang, W.~Zhou, W.~Pan, X.~Li, H.-T. Zhang, H.~Ding, and
  J.~Goncalves, ``Data driven discovery of cyber physical systems,''
  \emph{Nature communications}, vol.~10, no.~1, pp. 1--9, 2019.

\bibitem{SINDYc}
S.~L. Brunton, J.~L. Proctor, and J.~N. Kutz, ``Sparse identification of
  nonlinear dynamics with control (sindyc),'' \emph{IFAC-PapersOnLine},
  vol.~49, no.~18, pp. 710--715, 2016.

\bibitem{Implict-SINDy}
N.~M. Mangan, S.~L. Brunton, J.~L. Proctor, and J.~N. Kutz, ``Inferring
  biological networks by sparse identification of nonlinear dynamics,''
  \emph{IEEE Transactions on Molecular, Biological and Multi-Scale
  Communications}, vol.~2, no.~1, pp. 52--63, 2016.

\bibitem{Abrupt-SINDy}
M.~Quade, M.~Abel, J.~Nathan~Kutz, and S.~L. Brunton, ``Sparse identification
  of nonlinear dynamics for rapid model recovery,'' \emph{Chaos: An
  Interdisciplinary Journal of Nonlinear Science}, vol.~28, no.~6, p. 063116,
  2018.

\bibitem{two_step_method}
T.-T. Gao and G.~Yan, ``Autonomous inference of complex network dynamics from
  incomplete and noisy data,'' \emph{Nature Computational Science}, vol.~2,
  no.~3, pp. 160--168, 2022.

\bibitem{delemma_1}
D.~Feller, ``The role of databases in support of computational chemistry
  calculations,'' \emph{Journal of computational chemistry}, vol.~17, no.~13,
  pp. 1571--1586, 1996.

\bibitem{NN_advantage}
A.~A. Bataineh and D.~Kaur, ``A comparative study of different curve fitting
  algorithms in artificial neural network using housing dataset,'' in
  \emph{NAECON 2018 - IEEE National Aerospace and Electronics Conference},
  2018, pp. 174--178.

\bibitem{time_series_period}
M.~G. Elfeky, W.~G. Aref, and A.~K. Elmagarmid, ``Periodicity detection in time
  series databases,'' \emph{IEEE Transactions on Knowledge and Data
  Engineering}, vol.~17, no.~7, pp. 875--887, 2005.

\bibitem{DFT}
T.~Lu, Y.-F. Chen, B.~Hechtman, T.~Wang, and J.~Anderson, ``Large-scale
  discrete fourier transform on tpus,'' \emph{IEEE Access}, vol.~9, pp.
  93\,422--93\,432, 2021.

\bibitem{period-frequency}
Y.~Wang, \emph{Time-frequency analysis of seismic signals}.\hskip 1em plus
  0.5em minus 0.4em\relax John Wiley \& Sons, 2022.

\bibitem{NN_without_interablilty}
R.~Agarwal, L.~Melnick, N.~Frosst, X.~Zhang, B.~Lengerich, R.~Caruana, and
  G.~E. Hinton, ``Neural additive models: Interpretable machine learning with
  neural nets,'' \emph{Advances in Neural Information Processing Systems},
  vol.~34, pp. 4699--4711, 2021.

\bibitem{Weierstrass}
D.~P{\'e}rez and Y.~Quintana, ``A survey on the weierstrass approximation
  theorem,'' \emph{arXiv preprint math/0611038}, 2006.

\bibitem{dimensional_disaster}
Y.~Mao and F.~Ding, ``A novel parameter separation based identification
  algorithm for hammerstein systems,'' \emph{Applied Mathematics Letters},
  vol.~60, pp. 21--27, 2016.

\bibitem{NP_hard}
A.~Hernandez, F.~Ruiz, A.~Desideri, C.~Ionescu, S.~Quoilin, V.~Lemort, and
  R.~De~Keyser, ``Nonlinear identification and control of organic rankine cycle
  systems using sparse polynomial models,'' in \emph{2016 IEEE Conference on
  Control Applications (CCA)}.\hskip 1em plus 0.5em minus 0.4em\relax IEEE,
  2016, pp. 1012--1017.

\bibitem{RGA}
C.~Novara, ``Sparse identification of nonlinear functions and parametric set
  membership optimality analysis,'' \emph{IEEE Transactions on automatic
  control}, vol.~57, no.~12, pp. 3236--3241, 2012.

\bibitem{Least_angle_regression}
B.~Efron, T.~Hastie, I.~Johnstone, and R.~Tibshirani, ``Least angle
  regression,'' \emph{The Annals of statistics}, vol.~32, no.~2, pp. 407--499,
  2004.

\bibitem{improved_SBL_1}
J.~Jin, Y.~Yuan, W.~Pan, C.~Tomlin, A.~A. Webb, and J.~Gon{\c{c}}alves,
  ``Identification of nonlinear sparse networks using sparse bayesian
  learning,'' in \emph{2017 IEEE 56th Annual Conference on Decision and Control
  (CDC)}.\hskip 1em plus 0.5em minus 0.4em\relax IEEE, 2017, pp. 6481--6486.

\bibitem{improved_SBL_2}
W.~R. Jacobs, T.~Baldacchino, T.~Dodd, and S.~R. Anderson, ``Sparse bayesian
  nonlinear system identification using variational inference,'' \emph{IEEE
  Transactions on Automatic Control}, vol.~63, no.~12, pp. 4172--4187, 2018.

\bibitem{improved_SBL_3}
A.~Shahmansoori, ``Sparse bayesian multi-task learning of time-varying massive
  mimo channels with dynamic filtering,'' \emph{IEEE Wireless Communications
  Letters}, vol.~9, no.~6, pp. 871--874, 2020.

\bibitem{Runge}
Y.~Wu and M.~Zhu, ``Attitude reconstruction from inertial measurement:
  mitigating runge effect for dynamic applications,'' \emph{IEEE Transactions
  on Aerospace and Electronic Systems}, 2021.

\bibitem{water_conservancy}
``Noaa national water model conus retrospective dataset,''
  \url{https://registry.opendata.aws/nwm-archive/}.

\bibitem{temperature}
``Ncep-ncar reanalysis,'' https://psl.noaa.gov/data/gridded/data\newline
  .ncep.reanalysis.html.

\bibitem{Finance_data}
https://github.com/cccs-data/SIABF-and-Finance-data.git.

\bibitem{ResNet}
K.~He, X.~Zhang, S.~Ren, and J.~Sun, ``Deep residual learning for image
  recognition,'' in \emph{Proceedings of the IEEE conference on computer vision
  and pattern recognition}, 2016, pp. 770--778.

\bibitem{water_period}
P.~Hellard, M.~Avalos, F.~Guimaraes, J.-F. Toussaint, and D.~B. Pyne,
  ``Training-related risk of common illnesses in elite swimmers over a 4-yr
  period.'' \emph{Medicine and science in sports and exercise}, vol.~47, no.~4,
  pp. 698--707, 2015.

\end{thebibliography}
%
%
%
%
%
%
%
%
%
%




	
\end{document}